\pdfoutput=1

\documentclass[11pt]{article}

\usepackage{ACL2023}
\usepackage{url}

\usepackage{times}
\usepackage{latexsym}
\usepackage{listings}
\lstdefinestyle{json}{
    basicstyle=\ttfamily\small,
    breaklines=true,
    frame=none,
    numbers=none,
    numberstyle=\tiny,
    showstringspaces=false,
    tabsize=2,
    breakatwhitespace=false,
}
\lstdefinestyle{verbalized}{
    basicstyle=\ttfamily\small,
    breaklines=true,
    frame=none,
    numbers=none,
    showstringspaces=false,
    breakatwhitespace=false,
    breakindent=0pt,
}

\usepackage[T1]{fontenc}

\usepackage[utf8]{inputenc}

\usepackage{microtype}

\usepackage{inconsolata}
\usepackage{graphicx}

\title{Estimating Contamination via Perplexity: \\ Quantifying Memorisation in Language Model Evaluation}

\author{Yucheng Li \\
  University of Surrey, UK \\
  \texttt{yucheng.li@surrey.ac.uk}}

\begin{document}
\maketitle
\begin{abstract}
Data contamination in model evaluation is getting increasingly prevalent as the massive training corpora of large language models often unintentionally include benchmark samples. Therefore, contamination analysis has became an inevitable part of reliable model evaluation. However, existing method of contamination analysis requires the access of the entire training data which is often confidential for recent models. This prevent the community to rigorously audit these models and conduct accurate assessment of their capability. In this paper, we propose a novel method to quantify contamination without the access of the full training set, that measure the extent of contamination with perplexity. Our analysis provides evidence of significant memorisation of recent foundation models in popular reading comprehension, summarisation benchmarks, while multiple choice appears less contaminated.
\end{abstract}

\section{Introduction}

Recent years have seen remarkable progress in language models pre-trained on massive text corpora scraped from the web. However, many widely used evaluation benchmarks are also constructed from similar web sources, leading to a concerning issue of \textit{data contamination} where examples from test sets are unintentionally included in training data. Contamination enables models to "cheat" via memorisation of test data rather than displaying true generalisation \cite{marie2023}, which creates an illusion of progress, distorts model comparisons, and undermines the utility of benchmarks \cite{jacovi2023stop}.

Recent evaluation of language models usually involves a detailed contamination analysis of the benchmarks used \cite{brown2020language,chowdhery2022palm,touvron2023llama,openai2023gpt4}. These contamination reports typically contain two steps: 1) quantify potential test contamination by measuring n-gram overlap between the test set and training data, 2) compare model performance on clean vs. contaminated subsets. This procedure is essential for determining the validity of the evaluation procedure and the credibility of benchmarks in assessing model performance.

However, such analysis method relies on the access to the full training corpora, which are often unavailable for recent closed and open-sourced foundation models \cite{openai2023gpt4,touvron2023llama2}. Many fine-tuned language models released by the community also do not include reliable contamination reports and their fine-tuning datasets are not public either. This eliminates any possibility for the community to rigorously audit these models for contamination, which prevents reliable evaluation and accurate assessment of their capabilities. In addition, current method to identify potential test contamination, i.e., finding n-gram overlap between test set and training data, is also quite computation intensive, considering the massive size of modern training corpus.

\begin{figure}[t]
    \centering
    \includegraphics[width=\columnwidth]{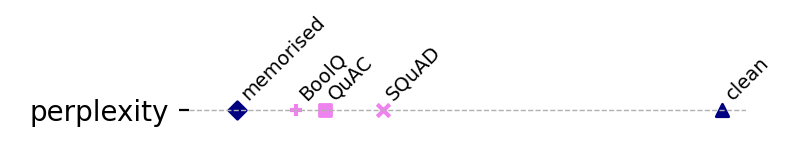}
    \caption{Perplexity comparison of three reading comprehension benchmarks against the \textit{memorised} and \textit{clean} baselines. \textsc{llama-30b} is the base model here.}
    \label{fig:single}
\end{figure}

In this paper, we propose a novel approach to quantify potential contamination in language model evaluation benchmarks without accessing the entire training data. Instead of identifying n-gram overlaps between training and test set, we directly observe whether models exhibit memorisation behaviour on test instances. \citet{carlini2021extracting,carlini2022quantifying} have defined the ``memorisation" of models that a sequence is considered as memorised if the model has considerably smaller perplexity on that sequence. The idea is that sequences leak in the training data will tend to have lower perplexity (i.e., higher likelihood) than sequences models never seen before. But simply computing the perplexity of evaluation benchmarks tells us nothing. We compare the perplexity of test set to two baselines: materials from the training data as the \textit{memorised} baseline, and materials completely not included in the training data as the \textit{clean} baseline. If the perplexity of test set is more close to the memorised baseline, then we have identified that the model exhibits significant memorisation on the test set, vice versa.

To make the comparison of perplexity fair, we ensure the baselines share the same source, format and length of the test set. For example, read comprehension benchmarks such as SQuAD, QuAD, and BoolQ employ Wikipedia articles as the context in their questions. Then when quantifying the contamination of the above benchmarks, we use Wikipedia as our baselines correspondingly. In Figure \ref{fig:single}, we illustrate how perplexity reflects the memorisation extent of model on Wikipedia.

Our proposed method enables contamination analysis for models without the need of their training data, which enables the community to conduct contamination analysis and perform credible evaluation. In addition, it avoids the computing of n-gram overlapping on massive training corpus. We formulate our method in Section 3 and present case studies for several typical language model evaluation benchmarks with our proposed method in Section 4. Our code and data can be found here: \url{https://github.com/liyucheng09/Contamination_Detector}.

\section{Background}

\noindent\textbf{What is data contamination?} Data contamination refers to the phenomenon that examples from the evaluation set are also found in the training data. This might lead to the evaluation failing to accurately reflect models' capabilities, as models can cheat by memorising instead of learning to generalise. There are two primary types of data contamination \cite{dodge2021documenting}: \textit{input contamination} refers to only the input appearing in the pretraining corpus, and \textit{input-and-label contamination} is when both inputs and their labels are present. The latter is generally more problematic, as models can directly memorise input-output pairs. But the first can still cause issues as models may gain an advantage by only input are learned, especially for assessing few-shot and zero-shot learning capabilities.

\noindent\textbf{How common is data contamination?} Data contamination appears to be quite widespread across commonly used NLP benchmark datasets based on findings from recent studies. \citet{dodge2021documenting} revealed exact match contamination rates ranging from under 2\% to over 50\% on various GLUE benchmarks when compared to the C4 pretraining data. The GPT-3 study \cite{brown2020language} found over 90\% of examples in Quac, SQuADv2, and DROP were flagged as contaminated. FLAN \cite{wei2021finetuned} evaluations identified 7 out of 26 datasets exhibiting a serious contamination ratio of 50\% and over. LLaMA 2 \cite{touvron2023llama} reported over 16\% of MMLU examples are contaminated and about 11\% are seriously contaminated (more than 80\% token leakage). GPT-4 \cite{openai2023gpt4} use academic exams instead of NLP benchmarks for model evaluation. While 4 out of 34 exams are found have zero contamination (e.g., Leetcode and Bar Exam), 9 out of 34 showed over 20\% of instances are marked as dirty examples. In summary, we found data contamination is becoming an increasingly prevalent issue as the popular of LLMs, which must be carefully measured and accounted for in order to accurately assess model performance.

\noindent\textbf{How to identify data contamination?} \citet{dodge2021documenting} takes a straightforward approach to detect exact matches between test set examples and the pretraining data after normalising for capitalisation and punctuation. The \textit{exact match here} means the entire input of an evaluation text is found in the training data. The GPT-3 paper \cite{brown2020language} uses n-gram overlap to identify contamination, treating any examples with 13-gram co-occurrence in both test sets and training data as dirty examples. PaLM \cite{chowdhery2022palm} considers a sample to be contaminated if 70\% of its 8-grams can be found at least once in the training data. LLaMA-2 matches on tokenized prompts and taking a bottom-up, token-level approach to identify contamination. Overall, existing approaches usually use substring matching between evaluation examples and training data to identify data contamination. However, if we have no access to the training data, which is often the case for most recent closed models, it is extreme difficult to reveal contamination by observing models themselves. Some pioneering studies \cite{carlini2021extracting,carlini2022quantifying} propose an approach to check whether a model has memorised an example by asking it to reconstruct the example verbatim, which can potentially used to detect evaluation data leakage.

\noindent\textbf{To what extent does data contamination affect model evaluation?} While contaminated data can potentially inflate scores, models do not necessarily perform worse on clean subsets or better on dirty subsets across all datasets. The degree of impact likely depends on factors like the dataset characteristics, model scale, and nature of the pre-training data. For instance, GPT-3 \cite{brown2020language} showed a small 1-2\% performance drop on clean subsets for PIQA and ReCoRD, comparing to a significant 6\% drop on clean set of SQuAD as 94\% of its test examples were contaminated. The LLaMA model \cite{touvron2023llama} did not show significant gaps between clean and dirty subset performance. On HellaSwag, LLaMA's 70B model showed a 15.3 point gap between clean (63.5) and dirty (78.8) subsets. Detecting and accounting for data contamination remains an active area of research, as there is no consensus yet on best methodologies and acceptable contamination levels.

\section{Method}

In this section, we explain our proposed method to quantify potential contamination of language model evaluation. Our method is comprised of three components: Perplexity Computing, Baseline Preparation, and Analysis.

\subsection{Perplexity Computing}
\label{perplexity}
In the case of a given benchmark dataset \( D \) and a language model \( M \), the first step involves verbalization of samples from \( D \). Specifically, we utilize a suitable prompt template to transform samples into an input sequence \( d \) that the model can process. To illustrate, consider a sample from the SQuAD \cite{rajpurkar2018know} dataset:

\begin{lstlisting}[style=json]
{
  "id": "56cf7d414df3c31400b0d849",
  "title": "Kanye_West",
  "context": "...",
  "question": "...",
  "answers": {
    "text": ...
  }
}
\end{lstlisting}
This sample is verbalised to fit a prompt designed for the SQuAD benchmark, resulting in:

\begin{quote}
"Title: Kanye West; Context: Myers spoke .....; Question: What happened after Kanye made his controversial statement? Answer: Rick Kaplan cut off the microphone and then cut away to Chris Tucker."
\end{quote}
The primary goal of this process is to adapt the sample so that it can be fed into the language model. Once the sample is verbalised, we compute its perplexity using the model \(M\) and the formula:

\[
P(d) = -\frac{1}{N} \sum_{i=1}^{N} \log_2 q(d_i|d_{<i})
\]
Here, \( N \) represents the length of the sequence \( d \) and \( q(d_i|d_{<i}) \) signifies the model's estimated conditional probability of \( d_i \) given the preceding tokens \( d_{<i} \). Note that the perplexity here refers to log-perplexity for better comparison. Lastly, the perplexities for individual sequences within \( D \) are averaged to arrive at a singular, representative perplexity value for the entire benchmark dataset.

\begin{figure*}
    \centering
    \includegraphics[width=\textwidth]{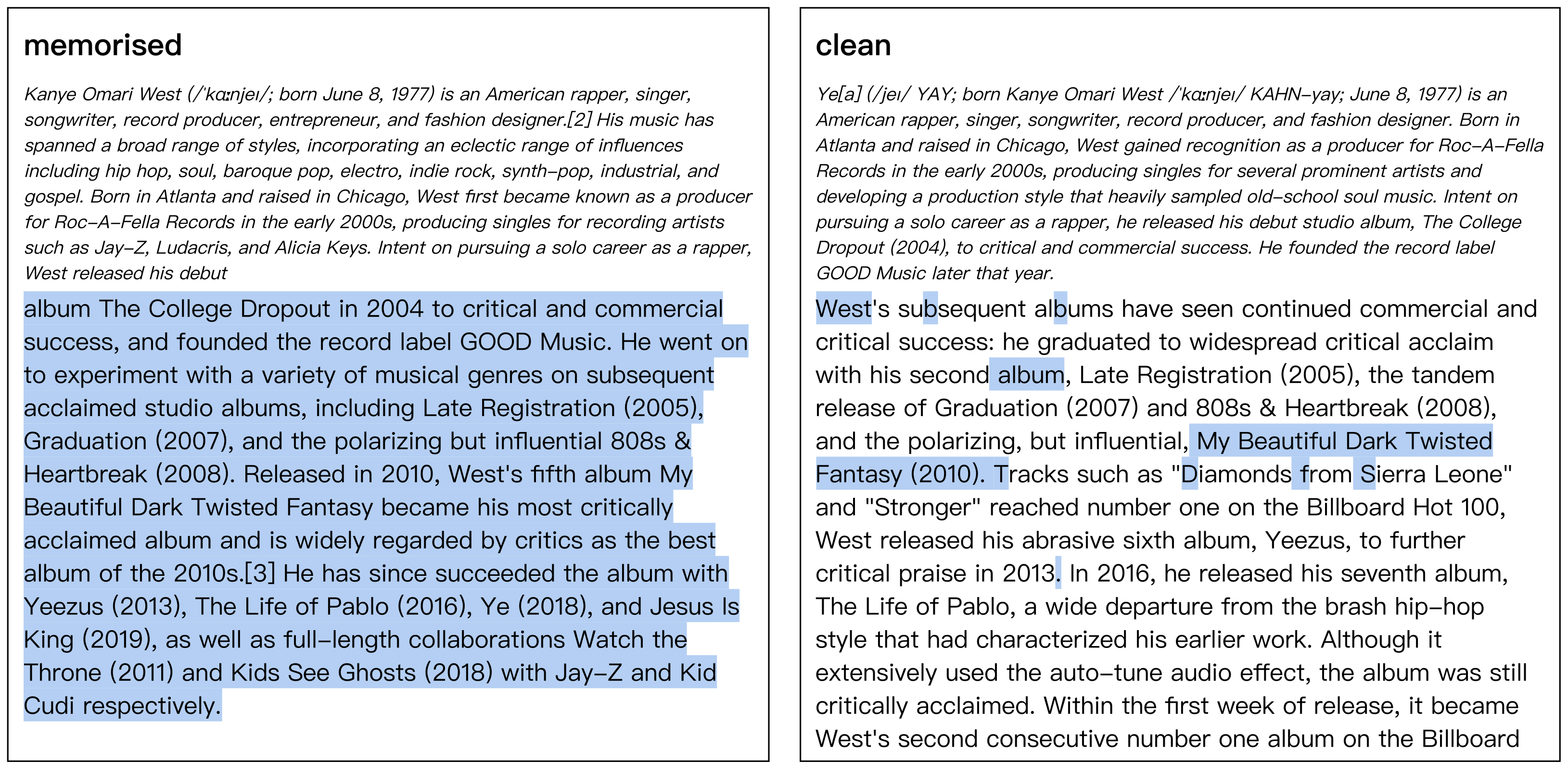}
    \caption{Visualisation of \textsc{gpt-3} memorisation on the \textit{memorised} and \textit{clean} baselines. The \textit{memorised} example is the 23 Dec 2019 version Wikipedia of Kanye West, and the \textit{clean} baseline is the same page in 27 Seq 2023.}
    \label{fig:string}
\end{figure*}

\subsection{Baseline Preparation}

The second step is the preparation of the two baselines for contamination comparison. The first is the \textit{memorised} baseline representing the perplexity of memorised materials for the language model \( M \). 
To establish this, we assemble texts that are likely to have been included in the model's training corpus. Based on the approximated training period, texts can be derived from time-specific snapshots of frequently used training datasets. For example, if the model \(M\) was trained on a Common Crawl dataset spanning 2016-2019, we would gather web pages specifically from this period. Similarly, if the model was trained on a 2019 dump of Wikipedia, we would collect articles from historical versions of Wikipedia corresponding to that year. The second is the \textit{clean} baseline representing the perplexity of fresh materials for \( M \). We collect texts that are 100\% fresh for the the language model, which can be texts created after the model was released. A comparison of the two baseline is illustrate in Figure \ref{fig:string}.

The nature of the texts in the two baselines is tailored to the specific characteristics of the benchmark \( D \) to ensure a fair perplexity comparison. But there are three main aspects that require special attention in preparing the baselines: the source, the format, and the length.
\begin{quote}
    \textbf{SQuAD\_v2}
    
    \quad Source: Wikipedia
    
    \quad Format: Reading Comprehension
    
    \quad Length: 127 words
\end{quote}
Let's take SQuAD benchmark as an example. We firstly notice the source of SQuAD is Wikipedia articles. Then our two baselines should employ the same source, that are the \textit{memorised} Wikipedia articles and \textit{fresh} Wikipedia articles. The former are Wikipedia articles in the training data of the language model $M$, the later are articles created after the language model $M$. The second to focus is the format. SQuAD is a reading comprehension benchmarks which means there are \textit{(passage, question, answer)} for each sample. We should make sure the two baselines share the same structure as the benchmark of interest. If our benchmark of interest is a multi-choice questions benchmark, here we should follow \textit{(questions, choices, answer)} format. At last, we need to ensure the baselines share the same length than the benchmark to obtain a fair perplexity comparison.

Followed by the preparation of the \textit{memorised} and \textit{clean} baselines, we compute perplexity of the two baselines as well as the benchmark itself as described in Section \ref{perplexity}: verbalisation and perplexity computing.

\subsection{Analysis}

The third and final step is to interpret the comparison results. Specifically, if the perplexity of \( D \) is closer to the \textit{memorised} Baseline, it suggests that the model exhibits a high degree of memorisation. Conversely, if it aligns more closely with the \textit{clean} Baseline, this would indicate that the level of memorisation is either low or negligible. We present case studies on popular language model benchmarks in the next section.

\begin{figure*}[th]
    \centering
    \includegraphics[width=0.83\textwidth]{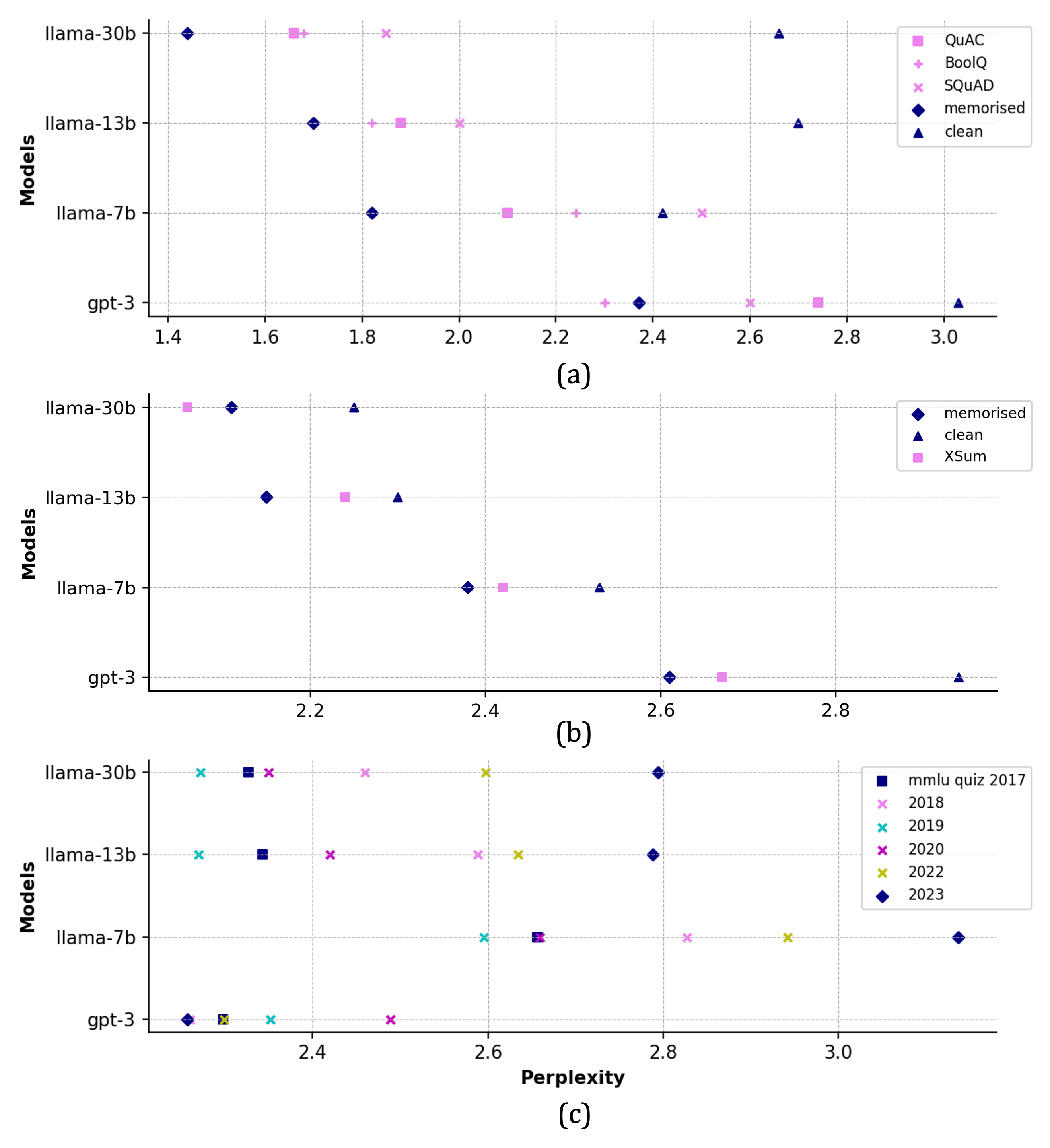}
    \caption{Perplexity comparison between benchmarks against the \textit{memorised} and \textit{clean} baselines.}
    \label{fig:case}
\end{figure*}

\section{Experiments}

Here we analyse three major types of language model benchmarks with our proposed method: reading comprehension, summarisation, and multi-choice questions. Benchmarks are tested on four foundation models: \textsc{gpt-3} \cite{brown2020language}, \textsc{llama-7b,13b,30b} \cite{touvron2023llama}.

\subsection{Reading Comprehension}

We analyse three popular reading comprehension benchmarks: QuAC \cite{choi2018quac}, BoolQ \cite{clark2019boolq}, and SQuAD\_v2. First, the above three benchmarks follow the standard reading comprehension format that each sample contains a \textit{(passage, question, answer)} tuple. We only include \textit{passage} in our contamination analysis, as we are unable to find available \textit{question} and \textit{answer} as clean baseline. Second, they all share the same source that the \textit{passages} are all comprised with Wikipedia articles. Therefore, we collect Wikipedia articles as the \textit{memorised} and \textit{clean} baselines. We use latest Wikipedia article created in June and July 2023 as the \textit{clean} baseline. As all foundation models in our experiment are released before June 2023, we ensure the clean baseline is not presented in the training phrase of models. We choose Wikipedia in different period as the \textit{memorised} baseline for different foundation models. \citet{brown2020language} report \textsc{gpt-3} is trained on CommonCrawl and Wikipedia dump covering 2016 to 2019. And \citet{touvron2023llama} report \textsc{Llama} series are trained on Wikipedia dump from June-August 2022. Therefore, we sample Wikipedia articles from corresponding period as the \textit{memorised} baseline. The averaged length of the three reading comprehension benchmarks are 107 words so we truncate our baseline to the same length. The comparison results is presented in Figure \ref{fig:case} (a).

We found all four foundation model exhibit significant memorisation behaviour on the three benchmarks. For large foundation models, e.g., \textsc{gpt-3} and \textsc{llama-13b,30b}, benchmarks are largely memorised which leads to lower perplexity than the clean baseline. Although smaller model show less memorisation than larger models, \textsc{llama-7b} still seems to have memorised benchmarks to some extent.

\subsection{Summarisation}

Here we conduct contamination analysis on the summarisation benchmark Xsum \cite{narayan2018don}. Each sample from summarisation benchmark has \textit{(document, summary)}. Both document and summary are included in our contamination test. Note that we use the verbalisation method mentioned in Section \ref{perplexity} to formalise samples to language model input. The source of XSum is BBC news. Therefore we collect BBC news articles as our \textit{memorised} and \textit{clean} baseline. The period of \textit{memorised} baseline collection is the same as the reading comprehension benchmarks. And we use latest BBC News articles created in June 2023 as the \textit{clean} baseline. We control the length of baselines to the averaged length of XSum which is 358 words.

The results are presented in Figure \ref{fig:case} (b). Again, we find significant memorisation behaviour for all foundation models on XSum. For \textsc{llama-30b}, we even obtain a lower perplexity of XSum compared to the memorised baseline.

\subsection{Multi-Choice}

Here we conduct contamination analysis on the multi-choice question benchmarks. We choose MMLU \cite{hendrycks2021measuring} as the example. Due to the massive size of MMLU, we only analyse a small portion of quizzes from MMLU. Check out the quizzes we include in the Appendix \ref{mmlu}. Most quizzes are in Pdf format. We use Adobe to transfer Pdf files to .docx files then extract text from them. We measure the perplexity for quizzes ranging from 2017 to 2023 and present in Figure \ref{fig:case} (c). Note that verbalisation method in Section \ref{perplexity} is used to format the quizzes to language model input.

Based on the results, we don't identify clear contamination evidence for the multi-choices quizzes analysed. The perplexities has no observable correlation with the time distribution. Therefore we could say that there is no significant contamination risk for these quizzes we tested from MMLU. This could because most quizzes are from Pdf files and were not included in model training.

\section{Conclusion}

We propose a perplexity-based method to estimate language model benchmark contamination without full training data access. By comparing test perplexity to memorised and clean baselines, we can identify potential memorisation indicating leaked examples. Experiments on reading comprehension, summarisation, and multiple choice benchmarks demonstrate the approach can effectively detect contamination. This enables crucial contamination analysis when training data is unavailable, supporting credible evaluation.

\bibliography{custom}
\bibliographystyle{acl_natbib}

\appendix
\section{Multi-choice Quizzes}
\label{mmlu}
We list all quizzes we include in our multi-choice experiment here. All sources are from MMLU github repo.

\begin{itemize}
    \item \url{https://css.csail.mit.edu/}
    \item \url{https://www.engageny.org/resource/released-2019-3-8-ela-and-mathematics-state-test-questions}
    \item \url{https://www2.tesc.edu/tecep_desc/COM-210.pdf}
    \item \url{https://iaac.space/docs/problems/2019/IAAC_Final_Round_2019.pdf}
    \item \url{http://www.exploredatabase.com/2020/05/machine-learning-multiple-choice-questions-with-answers-home.html}
\end{itemize}

\end{document}